%% file: main.tex
\documentclass[10pt,twocolumn,letterpaper]{article}

\usepackage{iccv}              
\usepackage[accsupp]{axessibility}

\input{preamble}
\definecolor{iccvblue}{rgb}{0.21,0.49,0.74}
\usepackage[pagebackref,breaklinks,colorlinks,allcolors=iccvblue]{hyperref}

\title{QR-LoRA: Efficient and Disentangled Fine-tuning via QR Decomposition for Customized Generation}

\input{authors}
\begin{document}

\twocolumn[{%
\renewcommand\twocolumn[1][]{#1}%
\maketitle
\begin{center}
\vspace{-0.2cm}
        \includegraphics[width=0.98\linewidth]{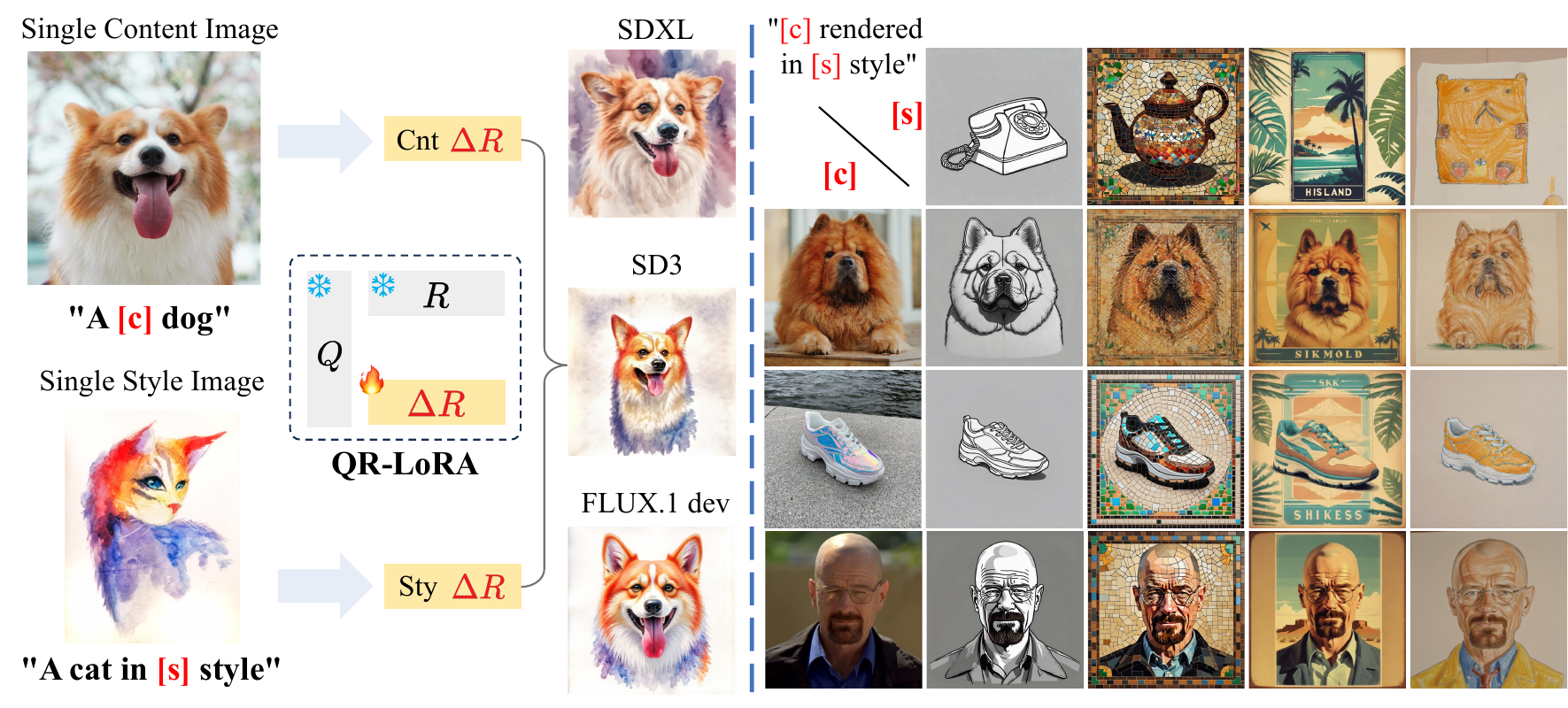}
        \captionof{figure}{Given a single content and a single style image, we present \textbf{\textit{QR-LoRA}}, a method to achieve efficient and disentangled control of content and style features through orthogonal decomposition. Our approach reduces trainable parameters while maintaining superior disentanglement properties, enabling flexible manipulation of visual attributes with enhanced initialization. 
        \textbf{\textit{Best viewed with zoom-in.}}
        } %
        \vspace{+0.2cm}
        \label{fig:teaser}
\end{center}
}]
\footnotetext[1]{Corresponding author}

\input{sec/0_abstract}

\input{sec/1_intro}

\input{sec/2_related}

\input{sec/3_method}

\input{sec/4_experiment}

\input{sec/5_discuss}

\section*{Acknowledgments}
This research is funded by the Chongqing Natural Science Foundation Innovation and Development Joint Fund (Changan Automobile) (Grant No. CSTB2024NSCQ-LZX0157), China.

{
    \small
    \bibliographystyle{ieeenat_fullname}
    \bibliography{main}
}



\end{document}

%% file: preamble.tex
\usepackage{times}
\usepackage{epsfig}
\usepackage{amsmath}
\usepackage{amssymb}
\usepackage[linesnumbered,ruled,vlined]{algorithm2e}
\usepackage{multibib}
\usepackage{xcolor,colortbl}
\usepackage{amsthm}

\usepackage{minitoc}
\usepackage{tocloft}

\usepackage{color}
\usepackage{booktabs}   
\usepackage{multirow}   
\usepackage{array}      
\usepackage{paralist}   
\usepackage{caption}    
\usepackage{tabularx}  
\definecolor{turquoise}{cmyk}{0.65,0,0.1,0.3}
\definecolor{purple}{rgb}{0.65,0,0.65}
\definecolor{dark_green}{rgb}{0, 0.5, 0}
\definecolor{orange}{rgb}{0.8, 0.6, 0.2}
\definecolor{red}{rgb}{0.8, 0.2, 0.2}
\definecolor{darkred}{rgb}{0.6, 0.1, 0.05}
\definecolor{blueish}{rgb}{0.3, 0.3, .6}
\definecolor{light_gray}{rgb}{0.7, 0.7, .7}
\definecolor{pink}{rgb}{1, 0, 1}
\definecolor{greyblue}{rgb}{0.25, 0.25, 1}
\definecolor{awesome}{rgb}{1.0, 0.13, 0.32}

\definecolor{figred}{rgb}{0.9, 0.1, 0.1}
\definecolor{figgreen}{rgb}{0.1, 0.7, 0.1}
\definecolor{figblue}{rgb}{0.1, 0.1, 0.9}
\definecolor{figmagenta}{rgb}{0.8, 0.1, 0.8}

\definecolor{reviewer7C2W}{RGB}{207, 49, 49} 
\definecolor{reviewerAT7d}{RGB}{34,139,34} 
\definecolor{reviewerPc7K}{RGB}{255,140,0} 

\usepackage{blindtext}

\renewcommand{\paragraph}[1]{\vspace{1em}\noindent\textbf{#1}}

\usepackage{gensymb}

\usepackage{graphicx}               

\usepackage{tabu}


%% file: authors.tex
\author{Jiahui Yang\textsuperscript{\rm 1,2}~~~Yongjia Ma\textsuperscript{\rm 2}~~~Donglin Di\textsuperscript{\rm 2}~~~Jianxun Cui\footnotemark[1]\textsuperscript{\rm ~~1} \\ Hao Li\textsuperscript{\rm 2}~~~Wei Chen\textsuperscript{\rm 2}~~~Yan Xie\textsuperscript{\rm 2}~~~Xun Yang\textsuperscript{\rm 3}~~~Wangmeng Zuo\textsuperscript{\rm 1} \\
\textsuperscript{\rm 1} Harbin Institute of Technology 
\hspace{0.3cm} \textsuperscript{\rm 2} Li Auto \hspace{0.3cm} 
\textsuperscript{\rm 3} University of Science and Technology of China
}

%% file: sec/0_abstract.tex
\begin{abstract}

Existing text-to-image models often rely on parameter fine-tuning techniques such as Low-Rank Adaptation (LoRA) to customize visual attributes. However, when combining multiple LoRA models for content-style fusion tasks, unstructured modifications of weight matrices often lead to undesired feature entanglement between content and style attributes.
We propose QR-LoRA, a novel fine-tuning framework leveraging QR decomposition for structured parameter updates that effectively separate visual attributes. Our key insight is that the orthogonal Q matrix naturally minimizes interference between different visual features, while the upper triangular R matrix efficiently encodes attribute-specific transformations.
Our approach fixes both Q and R matrices while only training an additional task-specific $\Delta R$ matrix. This structured design reduces trainable parameters to half of conventional LoRA methods and supports effective merging of multiple adaptations without cross-contamination due to the strong disentanglement properties between $\Delta R$ matrices.
Experiments demonstrate that QR-LoRA achieves superior disentanglement in content-style fusion tasks, establishing a new paradigm for parameter-efficient, disentangled fine-tuning in generative models.
The project page is available at: \url{https://luna-ai-lab.github.io/QR-LoRA/}
\end{abstract}

%% file: sec/1_intro.tex
\vspace{-0.25cm}
\section{Introduction}

Feature disentanglement and control in generative models have been a long-standing challenge, particularly in the domain of image synthesis where precise manipulation of visual attributes is crucial for practical applications.
Recent advances in diffusion models \cite{ho2020denoising,song2020denoising,peebles2023scalable,rombach2022high,song2020score,nichol2021improved,karras2022elucidating,zhang2023adding,ma2025tuningfreelongvideogeneration,ma2025adamsbashforthmoultonsolver,ye2023ip} have demonstrated remarkable capabilities in generating high-quality images. These models \cite{zhang2023adding,ye2023ip,wang2024instantstyle,jain2022zero,podell2023sdxl,flux2023,esser2024sd3,ruiz2023dreambooth,sohn2024styledrop,hertz2024style,jeong2024visual,jang2024decor} have shown impressive progress in both content generation and style manipulation, enabling diverse applications from photorealistic image synthesis to artistic style transfer. The rapid development of foundation models \cite{ho2020denoising,song2020score,song2020denoising,karras2022elucidating,10.1145/3404835.3462823} and adaptation techniques \cite{hu2021lora,ruiz2023dreambooth,mou2024t2i,zhang2023adding,ye2023ip} have made customizing generative models for specific visual tasks increasingly accessible, marking a milestone in AI-driven creative applications.

Contemporary approaches to feature disentanglement can be broadly categorized into content-specific \cite{ruiz2023dreambooth,zhang2023adding,mou2024t2i}, style-specific \cite{sohn2024styledrop,hertz2024style,wang2024instantstyle}, and joint content-style methods \cite{shah2025ziplora,frenkel2025blora,liu2024unziplora}. While single-task approaches have shown success in their respective domains, the increasing demand for simultaneous control over both content and style has led to the development of more comprehensive solutions. Among various model adaptation techniques, Low-Rank Adaptation (LoRA) \cite{hu2021lora} has emerged as a particularly promising approach for its parameter efficiency and versatility. Yet, current joint content-style methods based on LoRA face significant challenges in achieving effective feature disentanglement while maintaining parameter efficiency.
Recent solutions like ZipLoRA \cite{shah2025ziplora} attempt to achieve effective disentanglement by learning orthogonal coefficients between independently trained LoRAs, highlighting the importance of orthogonality in feature separation. However, such approaches, along with methods like B-LoRA \cite{frenkel2025blora}, ComposLoRA \cite{zhong2024multi}, CMLoRA \cite{zou2025cmlora}, remain heavily dependent on model-specific architectures \cite{rombach2022high,podell2023sdxl} and domain constraints. While the pursuit of orthogonality provides valuable insights for feature disentanglement, the field still lacks a unified, model-agnostic approach that can directly learn orthogonal adaptation modules without relying on post-training coefficient optimization. 

In this paper, we introduce QR-LoRA, a novel fine-tuning framework that leverages QR decomposition \cite{golub1996matrix} for disentangled content and style control. Our approach is motivated by both theoretical foundations and empirical observations. Through empirical analysis in Figure~\ref{fig:matrix-comparsion}, we observe that $Q$ matrices exhibit remarkably high similarity (approaching $1.0$) across different adaptation tasks, while $R$ matrices show relatively lower similarity. This stability of $Q$ matrices is theoretically supported by the minimal Frobenius norm property of orthogonal parameterization (detailed in Appendix), which ensures that the orthogonal basis vectors maintain consistent orthogonality while minimizing redundant transformations in the feature space. Based on these insights, we propose a novel initialization paradigm that maintains $Q$ as fixed orthogonal basis vectors spanning the linear transformation space, while introducing a $\Delta R$ mechanism to learn task-specific features, ensuring effective feature disentanglement while minimizing interference with the model's original capabilities.

Our framework begins with singular value decomposition (SVD)-based \cite{eckart1936approximation,golub1971singular,han2023svdiff,si2025lora-dash,meng2024pissa,lingam2024svft} core information extraction, followed by orthogonal basis construction through QR decomposition. The key innovation lies in our $\Delta R$ mechanism, which learns compact, task-specific transformations in the orthogonal space while preserving the model's original capabilities. This approach not only achieves more precise feature disentanglement but also significantly reduces the number of trainable parameters. 
Unlike existing methods that rely on model-specific architectural analysis \cite{frenkel2025blora} or complex merging strategies for independently trained LoRAs \cite{shah2025ziplora}, our method provides an elegant, model-agnostic solution that maintains consistent performance across various architectures. Extensive experiments demonstrate that QR-LoRA achieves superior disentanglement capabilities and faster convergence compared to conventional approaches, while requiring only half of the trainable parameters.
Our main contributions are as follows:

\vspace{-0.15cm}
\begin{itemize}
    \item A novel SVD-QR based initialization paradigm for LoRA that constructs orthogonal basis vectors through matrix decomposition, reducing trainable parameters by half while maintaining model performance.
    \vspace{-0.25cm}
    \item A residual mechanism ($\Delta R$) that learns task-specific transformations in a fixed orthogonal space, enabling precise control with disentanglement properties between visual attributes for effective merging.
    \vspace{-0.25cm}
    \item We demonstrate superior performance of QR-LoRA in content-style fusion tasks, with consistent results across various model architectures, highlighting its significant potential for broader applications.
\end{itemize}

%% file: sec/2_related.tex
\section{Related Work}
\label{sec:rel}

\textbf{Parameter-Efficient Fine-tuning (PEFT)} has emerged as a pivotal research area in large-scale generative models \cite{xu2023parameter,li2024vblora,ni2024pace}, particularly in adapting foundation models with minimal computational overhead. Various PEFT strategies have been proposed, including Adapter modules \cite{sung2022vl} that insert trainable layers, Prompt-tuning \cite{lester2021power} that optimizes continuous prompts, and LoRA \cite{hu2021lora} that decomposes weight updates into low-rank matrices (\ie, $AB$). Among these, LoRA has demonstrated remarkable success in large language models, inspiring innovative designs such as HydraLoRA \cite{tian2024hydralora} with asymmetric structures that leverage shared $A$ matrices for global feature integration and task-specific $B$ matrices for capturing intrinsic components, effectively minimizing interference between different tasks. Recent advances like PISSA \cite{meng2024pissa} leverage SVD-based initialization for accelerated convergence, while SVDiff \cite{han2023svdiff} introduces a compact parameter space by fine-tuning singular values of weight matrices, significantly reducing model size while maintaining performance. SVFT \cite{lingam2024svft} further advances the field by efficiently updating weights through sparse combinations of singular vectors, achieving remarkable performance with minimal trainable parameters. These advances in parameter-efficient learning have provided valuable insights for text-to-image model adaptation. However, the direct application of these techniques to visual generation tasks faces unique challenges due to the complex nature of visual feature disentanglement, especially in scenarios requiring fine-grained control over both content and style attributes.

\textbf{Image customization} in text-to-image models has evolved along several distinct technical trajectories. Model adaptation approaches \cite{sohn2024styledrop,ye2023ip,zhang2023adding,mou2024t2i,shah2025ziplora,liu2024unziplora,li2024vblora} encompass various strategies from parameter-efficient modules to feature injection mechanisms and lightweight model adjustments, enabling flexible customization through different architectural modifications. Content-preserving methods \cite{ruiz2023dreambooth,xiao2023fastcomposer,ye2023ip,kumari2022customdiffusion} emphasize maintaining subject identity and semantic consistency through specialized training strategies, introducing techniques for robust feature preservation and identity-aware generation. Advanced control mechanisms have emerged through attention-based techniques \cite{hertz2024style,jeong2024visual,alaluf2024cross,kumari2022customdiffusion,hertz2022prompt} and plug-and-play solutions \cite{tumanyan2023plug}, alongside lightweight training-free alternatives including editing-based approaches \cite{chen2023trainingfree,epstein2023diffusion,wang2024instantstyle,yang2023zero,pan2024finding,jang2024decor,farzad2025lorax,mokady2023null,gal2022image,lei2025stylestudio} that enable flexible attribute manipulation through direct image editing and feature-level modifications without model retraining.
Despite these developments, existing approaches face inherent limitations in achieving effective feature disentanglement while maintaining parameter efficiency, particularly when handling multiple customization objectives. These challenges stem from the complex nature of visual feature separation and the computational overhead required for precise attribute control, motivating our investigation into a more unified and parameter-efficient approach that leverages orthogonal basis vectors for disentanglement.

\textbf{Feature disentanglement}, particularly in the context of content and style separation, has been extensively studied in text-to-image models. Traditional approaches \cite{ruiz2023dreambooth,zhang2023adding,mou2024t2i} focus on single-aspect control but struggle with joint manipulation of multiple attributes.
Several methods have been proposed for merging multiple LoRA adaptations, such as Custom Diffusion \cite{kumari2022customdiffusion}, Mix-of-Show \cite{gu2023mixofshow}, Ortha \cite{po2024orthogonal}, LoRACLR \cite{LoRACLR}, they often require complex training strategies or additional regularization terms.
In parallel, model-specific architectures \cite{wu2023uncovering,shah2025ziplora,li2024vblora} partially address this limitation through specialized designs for feature disentanglement, though their reliance on specific architectural choices limits their generality, with methods like B-LoRA \cite{frenkel2025blora} leveraging SDXL's architectural characteristics to identify specific blocks crucial for style-content separation, and ZipLoRA \cite{shah2025ziplora} proposing efficient merging strategies for independently trained style and subject LoRAs. While these approaches demonstrate promising results within their targeted frameworks, they remain heavily dependent on model-specific architectures (particularly SDXL) and often face challenges in maintaining both subject and style fidelity during joint generation.

%% file: sec/3_method.tex
\vspace{+0.15cm}
\section{Method}
\label{sec:meth}

\subsection{Preliminaries}

\textbf{Continuous generative frameworks.}
Modern generative models often construct samples through learned continuous dynamics that bridge data distributions and tractable priors. Two prominent paradigms—diffusion models \cite{ho2020denoising,song2020score,rombach2022high,podell2023sdxl} and flow matching \cite{liu2022flow,lipman2022flow,esser2024sd3,flux2023}, share this philosophy but differ in their construction of probability paths and training objectives. Both frameworks leverage continuous transformations to map between a complex data distribution and a simple prior distribution, enabling efficient sampling and learning.
Diffusion models gradually corrupt observed data $\mathbf{x}$ by mixing it with Gaussian noise ($\boldsymbol{\epsilon}$) over time:
$\mathbf{z}_t=\alpha_t\mathbf{x}+\sigma_t\boldsymbol{\epsilon}$.
The goal is to learn a reverse process that can reconstruct the original data from the noisy samples.
Training involves estimating either $\hat{\mathbf{x}}$ or $\hat{\boldsymbol{\epsilon}}$ using a neural network:
\vspace{-0.15cm}
\begin{equation}\mathcal{L}(\mathbf{x})=\mathbb{E}_{t\sim\mathcal{U}(0,1),\boldsymbol{\epsilon}\sim\mathcal{N}(0,\mathbf{I})}\left[w(\lambda_t)\cdot\frac{\mathrm{d}\lambda}{\mathrm{d}t}\cdot\|\boldsymbol{\hat{\epsilon}}-\boldsymbol{\epsilon}\|_2^2\right],\label{eq:diff}\end{equation}
where $\lambda_t=\log(\alpha_t^2/\sigma_t^2)$ is the log signal-to-noise ratio, and $w(\lambda_t)$ is a weighting function that balances the importance of different noise levels \cite{kingma2024understanding}.
Flow matching views the forward process as a linear interpolation between the data $\mathbf{x}$ and a noise term $\boldsymbol{\epsilon}$: $\mathbf{z}_t=(1-t)\mathbf{x}+t\boldsymbol{\epsilon}$.  
The key idea is to learn a velocity field $\mathbf{u}_t$ that describes the continuous transformation from noise to data:
\vspace{-0.15cm}
\begin{equation}\mathcal{L}_{\mathrm{CFM}}(\mathbf{x})=\mathbb{E}_{t\sim\mathcal{U}(0,1),\boldsymbol{\epsilon}\sim\mathcal{N}(0,\mathbf{I})}\left[\|\hat{\mathbf{u}}-\mathbf{u}\|_2^2\right],\label{eq:flow}\end{equation}
where $\hat{\mathbf{u}}$ is the predicted velocity field, which can be expressed as a linear combination of $\hat{\boldsymbol{\epsilon}}$ and $\mathbf{z}_t$.
This objective can be rewritten in terms of the noise prediction, similar to Eq.~\ref{eq:diff}, with an appropriate weighting function.

\begin{figure*}[th]
    \centering
    \includegraphics[width=\linewidth]{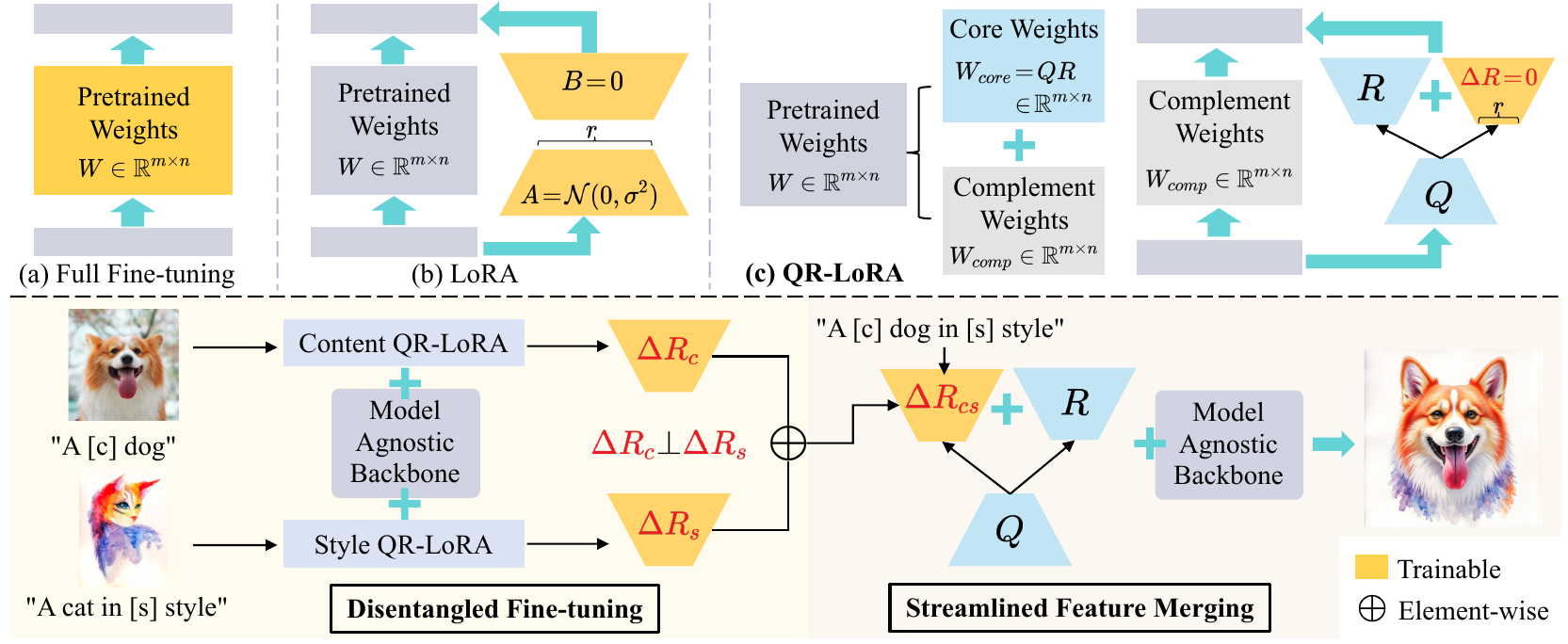}
    \caption{\textbf{Overview of QR-LoRA framework.} Upper (Sec~\ref{subsec:qr-lora}): Technical illustration of our method compared to traditional fine-tuning paradigms, highlighting our efficient parameter updates through orthogonal decomposition. 
    Lower (Sec~\ref{subsec:task-specific}): Application pipeline of our framework for content-style disentanglement, consisting of a disentangled fine-tuning module and a streamlined feature merging module.
    \vspace{-1.2em}
    }
    \label{fig:method}
\end{figure*}

\textbf{LoRA fine-tuning.}
LoRA has emerged as a parameter-efficient fine-tuning approach for adapting large pre-trained models to downstream tasks. Given a pre-trained weight matrix $W \in \mathbb{R}^{m \times n}$, LoRA learns a low-rank update $\Delta W = BA$, where $B \in \mathbb{R}^{m \times r}$ and $A \in \mathbb{R}^{r \times n}$ are trainable matrices with rank $r \ll \min(m,n)$. This decomposition significantly reduces the number of trainable parameters from $mn$ to $r(m+n)$. 
In text-to-image models \cite{podell2023sdxl,esser2024sd3,flux2023}, LoRA is typically injected into key components such as cross-attention and self-attention \cite{vaswani2017attention} layers, we provide detailed injection configurations in the Appendix.
During training, the original objective functions (\ie, Eq.~\ref{eq:diff} or Eq.~\ref{eq:flow}) remain unchanged, but the optimization is performed only on the low-rank parameters $A$ and $B$, while keeping the pretrained weights $W$ fixed, as shown in Figure~\ref{fig:method} (b).

\begin{figure}[t]
    \centering
    \includegraphics[width=\linewidth]{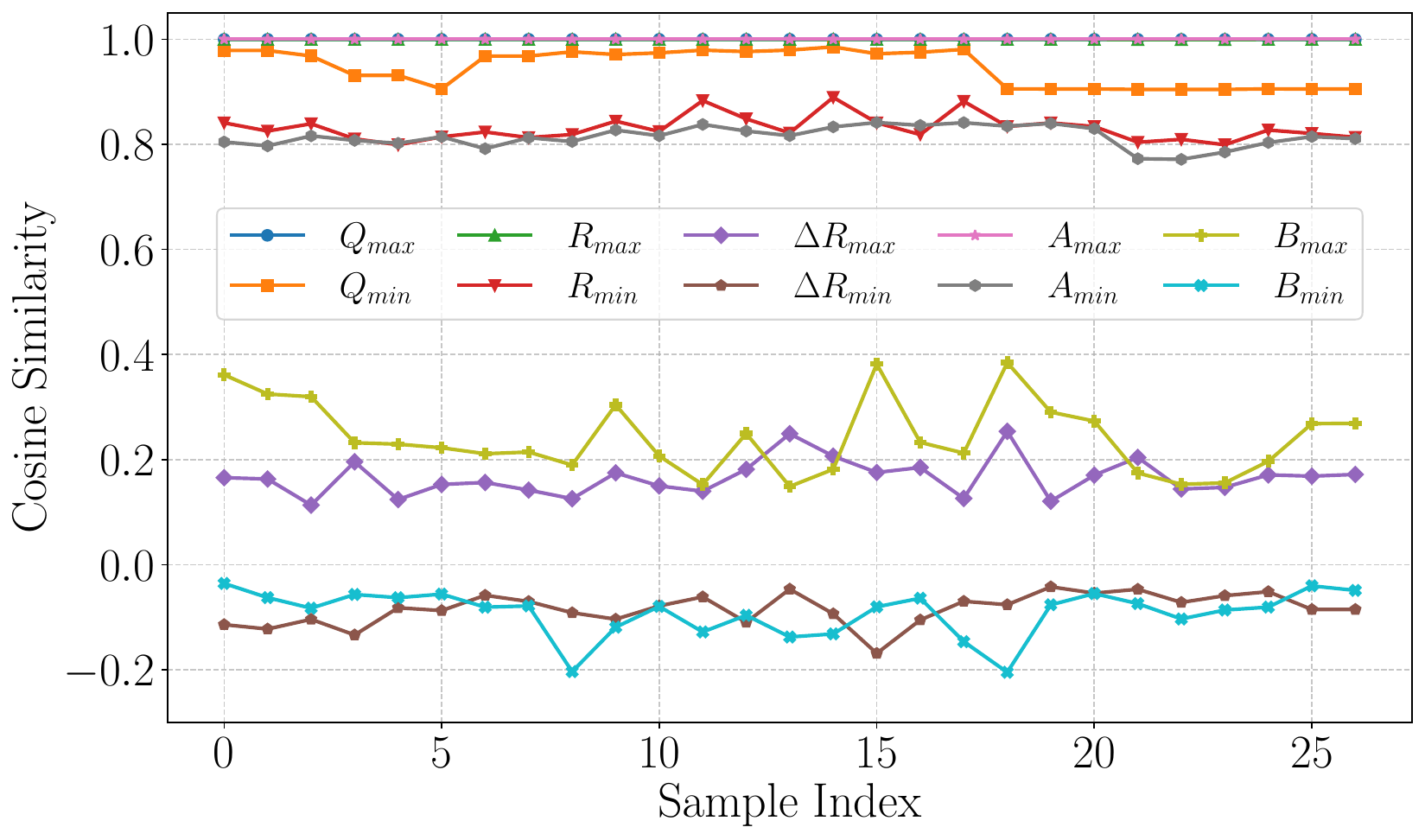}
    \caption{
    \textbf{Maximum and minimum cosine similarities across different matrices.} $Q$ and $R$ matrices are obtained from QR-LoRA framework with direct fine-tuning strategy, $\Delta R$ matrices from QR-LoRA framework with $\Delta R$-only update strategy, and $A$, $B$ matrices from vanilla LoRA decomposition. 
    The analysis reveals distinct similarity patterns across these matrix types, demonstrating their unique roles in feature representation. 
    All similarity values are computed between matrices trained on different image pairs.
    See Appendix for experimental cases and training details.
   \vspace{-0.2cm}
    }
    \label{fig:matrix-comparsion}
\end{figure}

\subsection{Motivation and Analysis}

\begin{figure}[t]
    \centering
    \includegraphics[width=\linewidth]{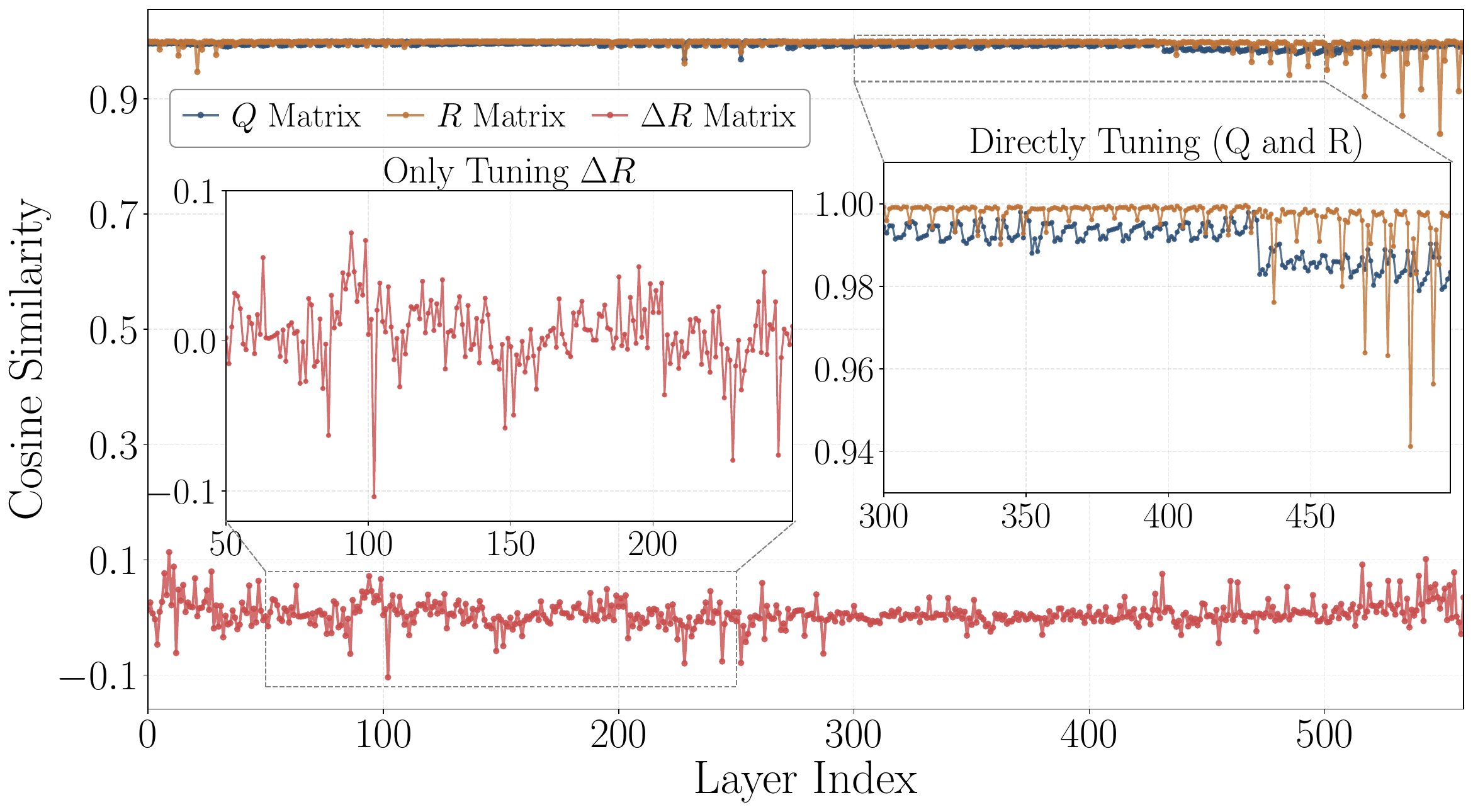}
    \caption{
    \textbf{Comparison of cosine similarities between different training strategies in QR-LoRA.} Visualization of layer-wise cosine similarities between matrices obtained from training on a randomly selected image pair in SDXL model. The comparison demonstrates distinct characteristics between directly fine-tuning $Q$ and $R$ matrices (upper zoom-in) versus only tuning $\Delta R$ matrices (lower zoom-in). See Appendix for comprehensive analyses across different model architectures and training data.
   \vspace{-0.3cm}
    }
    \label{fig:matrix-detail-comparsion-in-sdxl}
\end{figure}

Existing methods like ZipLoRA \cite{shah2025ziplora}, ComposLoRA \cite{zhong2024multi}, and CMLoRA \cite{zou2025cmlora}, while maintaining the original LoRA \cite{hu2021lora} training paradigm, share a common approach: they first train independent LoRA modules with the standard structure and then focus on studying their merging strategies. However, is this paradigm the optimal approach? This raises a fundamental question: \textbf{\textit{Could we redesign the LoRA structure from the ground up to achieve inherent feature disentanglement and better merging capabilities?}}
Through extensive matrix similarity analysis, we identify two critical insights that motivate our approach:

\textbf{Feature entanglement in direct weight updates.} The conventional LoRA training paradigm and direct QR decomposition training both exhibit similar feature entanglement issues. Our analysis reveals high cosine similarities between task-specific updates (Figure~\ref{fig:matrix-comparsion}), particularly in the matrices $A$ and $Q$. While the $B$ matrices show relatively lower similarities, the $R$ matrices demonstrate notably higher similarities ($>0.75$) due to our initialization strategy that inherits substantial prior information from the original weights (detailed in Algorithm~\ref{alg:qr_lora} and Sec~\ref{subsec:qr-lora}). This similarity pattern reveals a crucial insight: the stability of $Q$ matrices across different tasks suggests their potential as fixed orthogonal bases, while the $R$ matrices provides suitable flexibility for task-specific adaptations.
Since both approaches require training multiple components (\ie, $A$,$B$ and $Q$,$R$), achieving clean feature separation during merging becomes challenging. 
This observation motivates our design choice to focus on learning task-specific updates $\Delta R$ rather than training the entire decomposition. The effectiveness of this approach is validated by our empirical results in Figure~\ref{fig:matrix-comparsion} and Figure~\ref{fig:matrix-detail-comparsion-in-sdxl}, which show that the maximum cosine similarity between $\Delta R$ matrices across all injection layers remains consistently below 0.2, with mean values stable around 0, indicating superior feature disentanglement.

\textbf{Lack of structural constraints.} Most existing methods treat weight updates as unconstrained matrices, without leveraging mathematical properties that could facilitate feature disentanglement. Our investigation shows that imposing appropriate structural constraints through matrix decomposition can lead to more effective feature separation. Specifically, we find that orthogonal bases provide a natural framework for maintaining feature independence while allowing flexible combinations (see Appendix for detailed theoretical analysis).
Inspired by these observations and the properties revealed in Figure~\ref{fig:matrix-comparsion} and Figure~\ref{fig:matrix-detail-comparsion-in-sdxl}, we propose to leverage QR decomposition's unique characteristics. The orthogonal matrix $Q$ provides an ideal basis for representing features independently, while the upper triangular matrix $R$ captures the essential transformations.
This enables us to design a novel LoRA variant that naturally preserves feature disentanglement from the outset, rather than attempting to achieve it through post-training merging strategies.

\subsection{QR-LoRA Methodology}
\label{subsec:qr-lora}

Based on our analysis of the limitations in existing approaches and the insights gained from matrix similarity studies, we propose QR-LoRA, a novel parameter-efficient fine-tuning framework that naturally facilitates feature disentanglement through orthogonal decomposition. The key innovation of our approach lies in its unique initialization strategy and training mechanism, which we detail below.

\textbf{Core information extraction.} 
As illustrated in the upper part of Figure~\ref{fig:method}, compared to traditional full fine-tuning and LoRA, our QR-LoRA achieves more efficient parameter updates through orthogonal decomposition.
Given a pretrained weight matrix $W \in \mathbb{R}^{m \times n}$, we first employ SVD to systematically extract its core information structure (Figure~\ref{fig:method} (c)). This decomposition provides a natural way to identify and isolate the most significant components of the weight matrix:
\vspace{-0.1cm}
\begin{equation}\label{eq:svd}
W = U\Sigma V^\top= \sigma_1 u_1 v_1^\top + \sigma_2 u_2 v_2^\top + \dots + \sigma_s u_s v_s^\top,
\end{equation}
where $s\!=\!min(m,n)$, $U \in \mathbb{R}^{m \times s}$ contains the left singular vectors, $\Sigma \in \mathbb{R}^{s\times s}$ is a diagonal matrix of singular values arranged in descending order ($\sigma_{1}\geq\sigma_{2}\geq\cdots\geq\sigma_{s}$), reflecting the decreasing importance of feature dimensions, and $V^\top \in \mathbb{R}^{s \times n}$ contains the right singular vectors. By leveraging this natural ordering where larger singular values correspond to more significant feature transformations, we can construct a rank-$r$ core matrix that captures the most essential feature transformations:
\vspace{-0.1cm}
\begin{equation}\label{eq:w_core}
    W_{core} = U_{[:,:r]}\Sigma_{[:r]}V^\top_{[:r,:]} \triangleq \sigma_1 u_1 v_1^\top + \dots + \sigma_r u_r v_r^\top,
\end{equation}
this strategic truncation yields two complementary components: the core matrix $W_{core}$ that encapsulates the dominant feature transformations, and a complement matrix $W_{comp} = W - W_{core}$ that retains the residual information. This decomposition serves a dual purpose: it ensures that our subsequent QR-based adaptation focuses on the most significant components while maintaining structural consistency with the original LoRA framework through the initialization of the pretrained weight matrix $W$ with $W_{comp}$, which is outlined in Algorithm~\ref{alg:qr_lora} lines 1-4.

\textbf{Orthogonal basis construction.} 
To effectively capture and manipulate feature transformations, we leverage QR decomposition to construct orthogonal bases for our parameter updates. Rather than directly decomposing $W_{core}$, we strategically apply QR decomposition to its transpose $W_{core}^\top$, which aligns with our goal of maintaining consistent matrix multiplication order with the original LoRA formulation ($\Delta W = BA \Leftrightarrow R^\top Q^\top$). Given that $W_{core}^\top = V_{[:r,:]}\Sigma_{[:r]}U^\top_{[:,:r]}$, we can naturally express it as a product of two low-rank matrices ($S$ and $T$):
\begin{equation}
\begin{aligned}
    W_{core}^\top &= \underbrace{V_{[:r,:]}\Sigma_{[:r]}}_{S} \>\  \underbrace{U^\top_{[:,:r]}}_{T} = (Q_s R_s)T \\
    &= \underbrace{Q_s}_{Q} \underbrace{(R_s T)}_{R} \triangleq QR ,
\end{aligned}
\end{equation}
where $S$ characterizes the core structure of the column space of $W_{core}^\top$. By performing reduced QR decomposition ($Q_sR_s$) on $S$ rather than the full matrix, we obtain more precise orthogonal bases while maintaining computational efficiency. 
Through initializing the task-specific update matrix $\Delta R$ to zero, the final initialization of our QR-LoRA framework takes the form:
\vspace{-0.15cm}
\begin{equation}
    W_{origin}=W_{comp}+(Q(R+\Delta R))^\top,
\end{equation}
where $W_{origin}$ represents the original pretrained weights. By keeping both $Q$ and $R$ fixed during training, we provide stable anchors for task-specific adaptations through $\Delta R$. 
The complete process is detailed in Algorithm~\ref{alg:qr_lora}.

\subsection{Task-Specific Training} 
\label{subsec:task-specific}
As illustrated in the lower part of Figure~\ref{fig:method}, we apply our QR-LoRA framework to the challenging task of content-style disentanglement in image generation. Our approach consists of two key components that leverage the inherent orthogonality properties of our framework:

\textbf{Disentangled fine-tuning.} Through efficient fine-tuning on individual images, we separately learn content-specific ($\Delta R_c$) and style-specific ($\Delta R_s$) update matrices. The orthogonal basis $Q$ provided by our framework naturally facilitates this disentanglement by ensuring that different feature transformations remain independent (\ie, $\Delta R_c \perp \Delta R_s$ as shown in Figure~\ref{fig:matrix-comparsion}). This independence is further reinforced by our training strategy where each $\Delta R$ matrix is optimized for its specific task (\eg, content or style) while maintaining the shared orthogonal basis $Q$ and base matrix $R$ fixed. 

\textbf{Streamlined feature merging.} The inherent disentanglement properties of our learned $\Delta R$ matrices enable a remarkably simple yet effective merging strategy. Unlike existing approaches that require complex merging mechanisms, our method achieves flexible content-style integration through straightforward element-wise addition of the update matrices:
\vspace{-0.25cm}
\begin{equation}
    \Delta R_{cs} = \lambda_c \Delta R_c + \lambda_s \Delta R_s
\end{equation}
\vspace{-0.05cm}
where $\lambda_c$ and $\lambda_s$ are scaling coefficients that provide fine-grained control over the contribution of content and style features in the final generation. By default, $\lambda_c = \lambda_s = 1$.

\begin{algorithm}[t]
\SetKwInOut{To}{to}
\SetKwInput{KwIn}{Input}
\SetKwInput{KwOut}{Output}
\SetKwInOut{Initialization}{Initialization}
\DontPrintSemicolon
\caption{QR-LoRA for Efficient and Disentangled Fine-tuning with Orthogonal Basis}
\label{alg:qr_lora}

\KwIn{
Pretrained weight matrix $W \in \mathbb{R}^{m \times n}$;\\
Target task type $\tau$ (content or style);\\
Rank $r$ for low-rank approximation;
Learning rate $\eta$
}
\KwOut{
    Trained $\Delta R_{\tau}$;
    Orthogonal basis $Q$; \\
    Upper triangular base matrix $R$
}

\BlankLine
$U, \Sigma, V^\top \gets \textbf{SVD}(W)$ \hspace{0.5em} \tcp*{\textcolor{gray}{SVD of $W$}}
$W_{core} \gets U_{[:,:r]}\Sigma_{[:r]}V^\top_{[:r,:]}$ \hspace{0.5em} \tcp*{\textcolor{gray}{Core matrix}}
$W_{comp} \gets W - W_{core}; W_{origin}\gets W$ 

$W \gets W_{comp}$ \tcp*{\textcolor{gray}{Complement matrix}}
$S \gets V_{[:r,:]}\Sigma_{[:r]} \in \mathbb{R}^{n\times r}; T \gets U^\top_{[:,:r]} \in \mathbb{R}^{r\times m}$

$Q_s, R_s \gets \textbf{QR}(S); Q_s \in \mathbb{R}^{n\times r}, R_s \in \mathbb{R}^{r\times r}$
\tcp*{\textcolor{gray}{Reduced QR decomposition}}

$ Q \gets Q_s; R \gets R_s T \in \mathbb{R}^{r\times m}$

Initialize $\Delta R_{\tau} \gets 0_{r \times m}$ \hspace{0.5em} \tcp*{\textcolor{gray}{Init to zero}}

$W_{origin}=W_{comp}+(Q(R+\Delta R))^\top$

\While{not converged}{
    $\mathcal{L} \gets \textbf{TaskLoss}\{W_{comp} + (Q(R + \Delta R_{\tau}))^\top \}$    
    $\Delta R_{\tau} \gets \Delta R_{\tau} - \eta\nabla_{\Delta R_{\tau}}\mathcal{L}$ \hspace{0.5em} \tcp*{\textcolor{gray}{Update}}
}

\Return $\Delta R_{\tau}, Q, R$
\end{algorithm}

%% file: sec/4_experiment.tex
\section{Experiments}
\label{sec:exp}

\begin{figure*}[th]
    \centering
    \includegraphics[width=\linewidth]{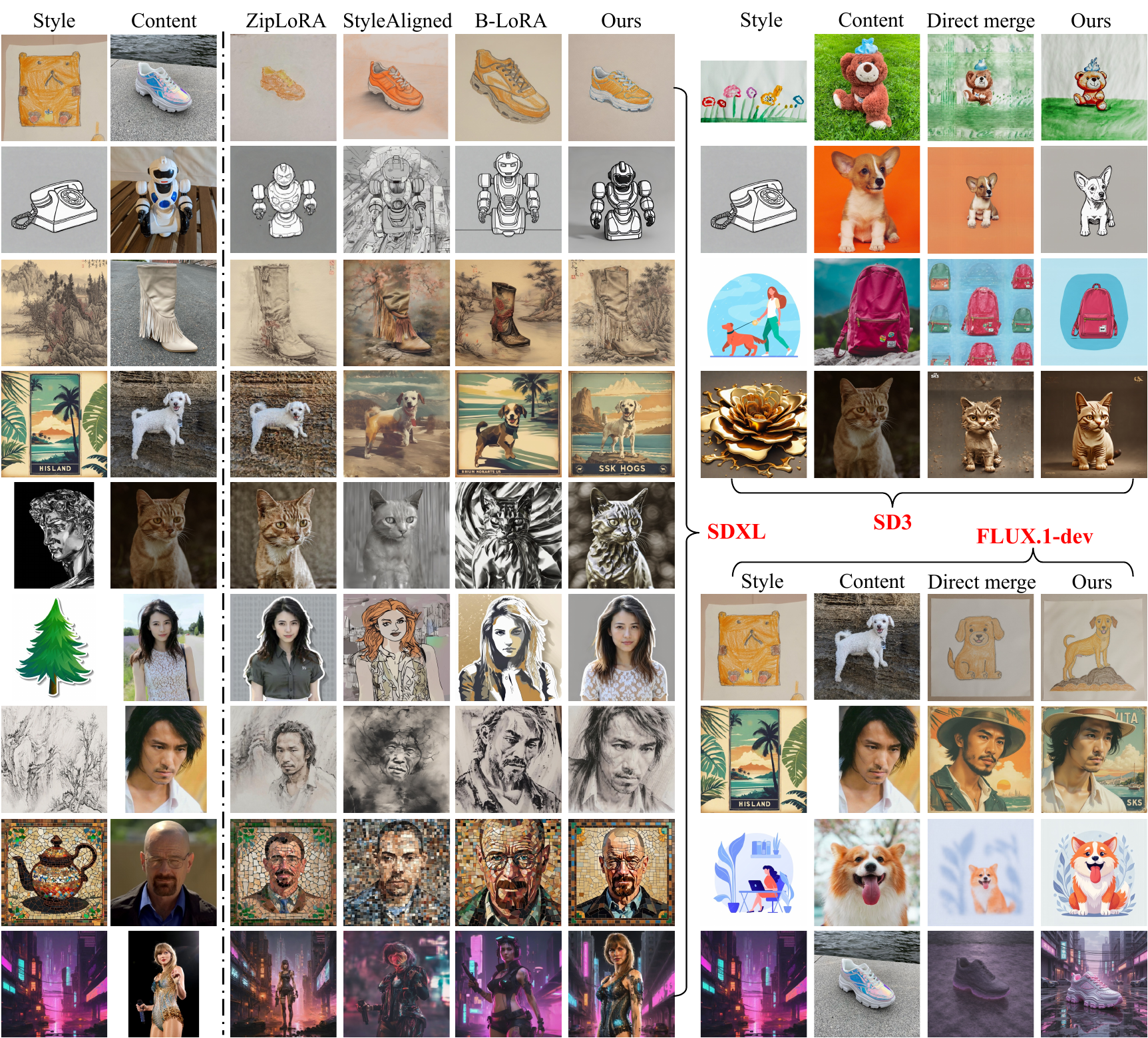}
    \caption{\textbf{Qualitative comparison.} Comparison of our QR-LoRA against state-of-the-art methods on SDXL and a naive baseline on SD3 and FLUX.1-dev models, demonstrating the model-agnostic nature and superior performance of our framework. Zoom in to view details.
    \vspace{-0.2cm}
    }
    \label{fig:comparison}
\end{figure*}

\begin{table*}[]
\centering
\caption{Quantitative comparison across different backbone models (SDXL, SD3, FLUX.1-dev). We report cosine similarities between generated images and their style (-S) or content (-C) references using DINO and CLIP features, along with user ratings (0-5 scale).}
\label{tab:quantitative-results}
\vspace{-1.5mm}
\resizebox{\textwidth}{!}{
\begin{tabular}{lcccccccccc}
\toprule
 & \multicolumn{4}{c}{SDXL \cite{podell2023sdxl}} & & \multicolumn{2}{c}{SD3 \cite{esser2024sd3}} & & \multicolumn{2}{c}{FLUX.1-dev \cite{flux2023}} \\ \cline{2-5} \cline{7-8} \cline{10-11} 
 & \rule{0pt}{10pt} ZipLoRA \cite{shah2025ziplora} & B-LoRA \cite{li2024vblora} & StyleAligned \cite{hertz2024style} & \textbf{Ours} & & DB-LoRA \cite{diffusers} & \textbf{Ours} & & DB-LoRA \cite{diffusers}& \textbf{Ours} \\ 
\midrule
DINO-S \cite{caron2021dino} $\uparrow$ & 0.686 $\pm$ 0.131 & 0.689 $\pm$ 0.075 & 0.651 $\pm$ 0.061 & \textbf{0.694} $\pm$ 0.084 & & 0.675 $\pm$ 0.043 & \textbf{0.711} $\pm$ 0.061 & & 0.723 $\pm$ 0.105 & \textbf{0.744} $\pm$ 0.077 \\
DINO-C \cite{caron2021dino} $\uparrow$ & 0.712 $\pm$ 0.112 & 0.740 $\pm$ 0.064 & 0.758 $\pm$ 0.053 & \textbf{0.776} $\pm$ 0.065 & & 0.673 $\pm$ 0.032 & \textbf{0.690} $\pm$ 0.040 & & 0.654 $\pm$ 0.072 & \textbf{0.687} $\pm$ 0.086 \\
CLIP-S \cite{radford2021clip} $\uparrow$ & 0.686 $\pm$ 0.105 & 0.669 $\pm$ 0.085 & 0.658 $\pm$ 0.076 & \textbf{0.707} $\pm$ 0.060 & & 0.652 $\pm$ 0.071 & \textbf{0.703} $\pm$ 0.056 & & 0.641 $\pm$ 0.074 & \textbf{0.690} $\pm$ 0.069 \\
CLIP-C \cite{radford2021clip} $\uparrow$ & 0.668 $\pm$ 0.149 & 0.646 $\pm$ 0.098 & 0.662 $\pm$ 0.114 & \textbf{0.709} $\pm$ 0.087 & & 0.771 $\pm$ 0.081 & \textbf{0.792} $\pm$ 0.077 & & 0.677 $\pm$ 0.125 & \textbf{0.709} $\pm$ 0.096 \\
User-Study $\uparrow$ & 3.13 & 3.34 & 3.67 & \textbf{4.07} & & 2.93 & \textbf{4.14} & & 2.86 & \textbf{3.96} \\
\bottomrule
\end{tabular}}
\vspace{-1.3em}
\end{table*}

\subsection{Implementation details}

\textbf{Experimental setup.}
To validate the model-agnostic nature of our framework, we evaluate QR-LoRA on multiple state-of-the-art diffusion models including SDXL \cite{podell2023sdxl}, SD3 \cite{esser2024sd3}, and FLUX.1-dev \cite{flux2023}. During the fine-tuning process, we keep the pretrained model weights and text encoders frozen. All experiments are conducted on a single image per training instance. We employ the Adam \cite{kingma2014adam} optimizer with a learning rate of $1e\text{-}4$ without data augmentations. The LoRA rank is set to $r=64$ with 500 training steps, requiring approximately 20 minutes on a single A800 GPU. Our training data is sourced from previous works \cite{ruiz2023dreambooth,sohn2024styledrop,xue2021end} and publicly available content.
Following previous works \cite{ruiz2023dreambooth,sohn2024styledrop,shah2025ziplora}, we adopt a standard prompt template for training. Specifically, we use $<\!\!c\!\!>$ to denote content-related trigger words (\eg, ``[c]" or ``sks") and $<\!\!s\!\!>$ for style-related expressions (\eg, ``[s]" or simple artistic description \cite{shah2025ziplora,li2024vblora}), combining them in the format ``a $<\!\!c\!\!>$ in $<\!\!s\!\!>$ style". 
For fair comparison, we maintain consistent content and style image settings across all comparative experiments.

\textbf{Compared methods.}
For SDXL \cite{podell2023sdxl}, where most recent style transfer methods are developed, we compare against state-of-the-art approaches including ZipLoRA \cite{shah2025ziplora}, StyleAligned \cite{hertz2024style}, and B-LoRA \cite{li2024vblora}. For SD3 \cite{esser2024sd3} and FLUX.1-dev \cite{flux2023} models, we compare against DreamBooth-LoRA \cite{diffusers} using naive weight combination.

\textbf{Metrics.}
To evaluate our method quantitatively, we compute cosine similarities between generated images and their references using DINO~\cite{caron2021dino} and CLIP~\cite{radford2021clip} features for both style (-S) and content (-C) aspects. We also conduct a user study to assess the perceptual quality, with detailed settings provided in the Appendix.

\begin{figure}[t]
    \centering
    \includegraphics[width=\linewidth]{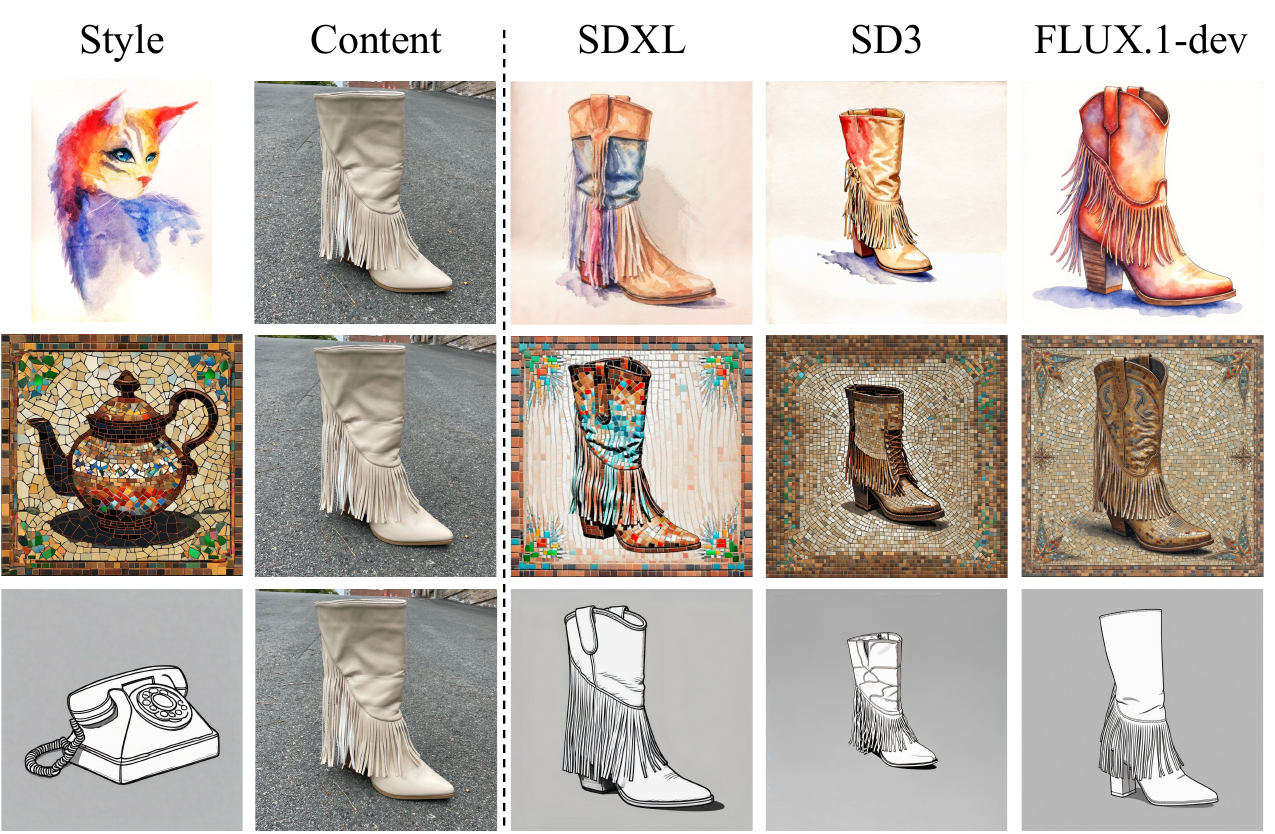}
    \caption{\textbf{Visualization on different backbone models.} Demonstration of QR-LoRA's generation results across different backbone models, showing consistent high-quality performance.}
    \label{fig:vis_diff_backbones}
    \vspace{-0.5em}
\end{figure}

\subsection{Results and Analysis}
We conduct comprehensive quantitative and qualitative experiments to evaluate our method, with additional experimental results provided in the supplementary material.

\textbf{Qualitative Analysis.}
Figure~\ref{fig:comparison} demonstrates the superior performance of QR-LoRA across various test scenarios. Our method consistently outperforms existing approaches in style transfer tasks, with particularly notable advantages in preserving human poses and anatomical details when handling complex body structures and dynamic postures.
Moreover, QR-LoRA exhibits strong model-agnostic capabilities, achieving consistent high-quality results across different backbone models including SDXL, SD3, and FLUX.1-dev, which is also demonstrated in Figure~\ref{fig:vis_diff_backbones}. This cross-model effectiveness demonstrates that our approach to feature manipulation and style transfer can be successfully applied to diverse model architectures while maintaining structural integrity of the generated images.

\textbf{Quantitative Analysis}
We evaluate our method on a test set of 64 randomly sampled generated images. As shown in Table~\ref{tab:quantitative-results}, QR-LoRA outperforms state-of-the-art methods on SDXL in both content preservation and style transfer, with significantly higher user ratings in our user study. On SD3 and FLUX.1-dev, our method consistently improves over the naive baseline DB-LoRA \cite{diffusers} (\ie, direct merge), demonstrating its model-agnostic effectiveness.

\begin{figure}[t]
    \centering
    \includegraphics[width=\linewidth]{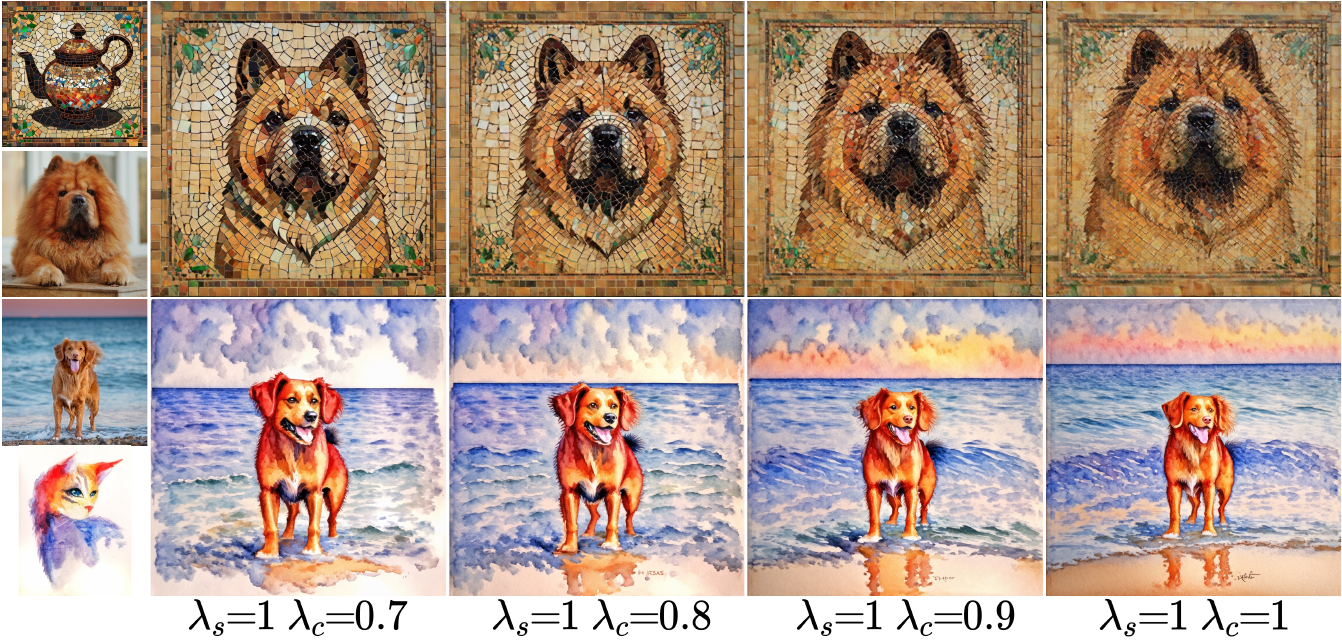}
    \caption{\textbf{Ablation study} on scaling coefficient $\lambda_c$ and $\lambda_s$.}
    \label{fig:ablation}
    \vspace{-0.9em}
\end{figure}

\textbf{Ablation.}
To further validate the robustness of our method, we experiment with different combinations of scaling coefficients $\lambda_c$ and $\lambda_s$ during feature composition. As shown in Figure~\ref{fig:ablation}, the consistent high-quality results across various coefficient pairs demonstrate that our approach is inherently robust, indicating that this scaling flexibility is a natural property rather than a requirement of our method.

\vspace{-0.3em}
\subsection{Application}

\begin{figure}[t]
    \centering
    \includegraphics[width=\linewidth]{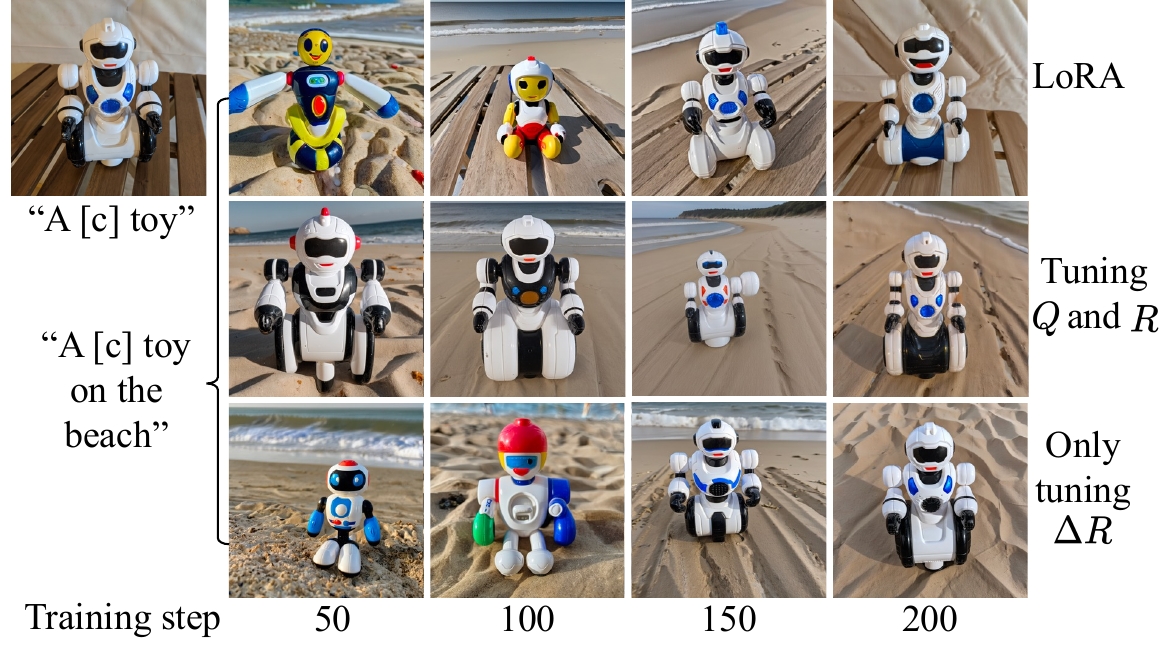}
    \caption{\textbf{Convergence analysis.} Comparison of training convergence between QR-LoRA and LoRA.
    \vspace{-0.1cm}
    }
    \label{fig:coverage_comparison}
\end{figure}

\textbf{Adaptive Training Strategies.}
Our framework provides flexible training configurations for different scenarios as shown in Figure~\ref{fig:coverage_comparison}. For feature disentanglement tasks, training only the $\Delta R$ matrices achieves comparable convergence to standard LoRA with better generalization. When faster convergence is needed, training both $Q$ and $R$ components simultaneously accelerates the process by leveraging pretrained structural information.

\begin{figure}[t]
    \centering
    \includegraphics[width=\linewidth]{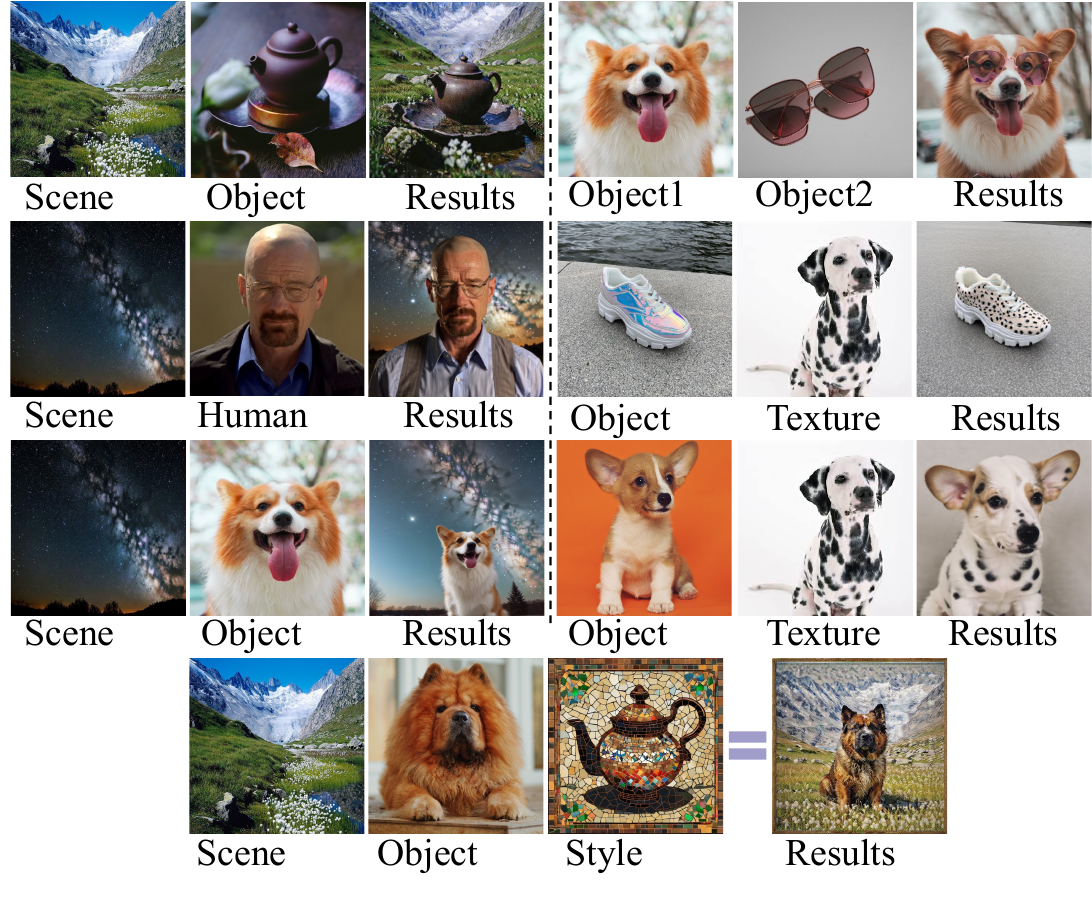}
    \caption{\textbf{Applications on multi-feature composition.} Our method enables flexible combination of various image attributes, demonstrating its capability beyond style-content synthesis.
    \vspace{-0.5cm}
    }
    \label{fig:applications}
\end{figure}

\textbf{Versatile Feature Composition.}
Beyond style-content synthesis, our method demonstrates remarkable capability in disentangled composition of various image attributes. As shown in Figure~\ref{fig:applications}, QR-LoRA effectively handles diverse feature combinations. This versatility highlights the method's potential for broader applications.

%% file: sec/5_discuss.tex
\section{Discussion}
\vspace{-0.6em}

\textbf{Limitation.}
While our method demonstrates strong performance, the incorporation of matrix decomposition operations during initialization introduces additional computational overhead. Specifically, this initialization process requires $\sim$1, 2, and 5 minutes for SDXL, SD3, and FLUX.1-dev respectively. However, this one-time cost during initialization does not affect subsequent inference efficiency.

\textbf{Future Work.}
The model-agnostic nature of QR-LoRA enables its extension to other modalities such as 3D \cite{raj2023dreambooth3d,yang2024tv} and video \cite{yang2024cogvideox,genmo2024mochi,9677948,yang2024robust,Zhou_2025_CVPR} customization.
Furthermore, our framework naturally aligns with Mixture-of-Experts (MoE) \cite{liu2024adamole} architectures when extended to multiple $\Delta R$ matrices, where studying the collaborative dynamics could enable more flexible and powerful control capabilities.